\documentclass{preprint}
\usepackage{booktabs}
\usepackage{graphicx}
\usepackage[colorlinks=true]{hyperref}
\usepackage[numbers,comma,sort&compress]{natbib} 
\usepackage{xcolor}
\usepackage{array}
\usepackage{mdframed}
\usepackage{enumitem}
\usepackage{tabularray}
\usepackage{xspace}
\usepackage{listings}
\usepackage{makecell}

\definecolor{Stone0}{HTML}{F6F3EB}
\newcommand\boxedname{Box\xspace} 
\newcounter{boxedcounter}
\newenvironment{pboxed}[1][]{
    \refstepcounter{boxedcounter}
    \begin{mdframed}[
        innertopmargin=1em, 
        innerbottommargin=2pt,
        innerleftmargin=2pt, 
        innerrightmargin=2pt,
        frametitleaboveskip=2pt,
        frametitlebelowskip=2pt,
        frametitle={\textbf{\boxedname \theboxedcounter:} #1},
        frametitlefont=\normalfont,
        frametitlerule=true,
        backgroundcolor=Stone0]%
    \setlength{\parindent}{0pt}%
    \setlength{\parskip}{0pt}%
}
{%
    \par%
    \end{mdframed}%
}

\definecolor{comment}{HTML}{333333}
\definecolor{string}{HTML}{008000}
\definecolor{number}{HTML}{098658}
\definecolor{key}{HTML}{0451A5}
\definecolor{bool}{HTML}{800080}
\definecolor{null}{HTML}{808080}
\definecolor{brace}{HTML}{000000}
\definecolor{comma}{HTML}{444444}
\lstdefinelanguage{json}{
    numberstyle=\scriptsize,
    stepnumber=1,
    showstringspaces=false,
    showspaces=false,
    showtabs=false,
    breaklines=true,
    basicstyle=\ttfamily\small,
    commentstyle=\color{comment}\ttfamily,
    stringstyle=\color{key}\ttfamily,
    keywordstyle=[1]\color{bool}\ttfamily,
    keywordstyle=[2]\color{null}\ttfamily,
    morestring=[b]",
    morecomment=[l]{//},
    morecomment=[s]{/*}{*/},
    keywords=[1]{true, false},
    keywords=[2]{null},
    upquote=false,
    literate=%
        {0}{{{\color{number}0}}}{1}
        {1}{{{\color{number}1}}}{1}
        {2}{{{\color{number}2}}}{1}
        {3}{{{\color{number}3}}}{1}
        {4}{{{\color{number}4}}}{1}
        {5}{{{\color{number}5}}}{1}
        {6}{{{\color{number}6}}}{1}
        {7}{{{\color{number}7}}}{1}
        {8}{{{\color{number}8}}}{1}
        {9}{{{\color{number}9}}}{1}
        {:}{{{\color{comma}:}}}{1}
        {,}{{{\color{comma},}}}{1}
        {\{}{{{\color{brace}\{}}}{1}
        {\}}{{{\color{brace}\}}}}{1}
        {[}{{{\color{brace}[}}}{1}
        {]}{{{\color{brace}]}}}{1}
}
\lstdefinelanguage{prompt}{
    breaklines=true,
    breakindent=10pt,
    basicstyle=\small\ttfamily,
    columns=fullflexible,
    upquote=false,
}

\title{RSNA Large Language Model Benchmark Dataset for Chest Radiographs of Cardiothoracic Disease: Radiologist Evaluation and Validation Enhanced by AI Labels (REVEAL-CXR)}\label{rsna-large-language-model-benchmark-dataset-for-chest-radiographs-of-cardiothoracic-disease-radiologist-evaluation-and-validation-enhanced-by-ai-labels-reveal-cxr}
\author[1,2]{Yishu Wei}
\author[3]{Adam E. Flanders}
\author[4]{Errol Colak}
\author[5]{John Mongan}
\author[6]{Luciano M Prevedello}
\author[7]{Po-Hao Chen}
\author[8]{Henrique Min Ho Lee}
\author[8]{Gilberto Szarf}
\author[8]{Hamilton Shoji}
\author[9]{Jason Sho}
\author[10]{Katherine Andriole}
\author[11]{Tessa Cook}
\author[12]{Lisa C. Adams}
\author[13]{Linda C. Chu}
\author[5]{Maggie Chung}
\author[1]{Geraldine Brusca-Augello}
\author[4]{Djeven P. Deva}
\author[14,15]{Navneet Singh}
\author[16]{Felipe {Sanchez Tijmes}}
\author[1]{Jeffrey B. Alpert}
\author[16]{Elsie T. Nguyen}
\author[11]{Drew A. Torigian}
\author[16]{Kate Hanneman}
\author[1]{Lauren K Groner}
\author[1]{Alexander Phan}
\author[17]{Ali Islam}
\author[4]{Matias F.Callejas}
\author[8]{Gustavo Borges da Silva Teles}
\author[12]{Faisal Jamal}
\author[9]{Maryam Vazirabad}
\author[15]{Ali Tejani}
\author[18]{Hari Trivedi}
\author[19]{Paulo Kuriki}
\author[19]{Rajesh Bhayana}
\author[1]{Elana T. Benishay}
\author[2]{Yi Lin}
\author[1,2]{Yifan Peng}
\author[1,*]{George Shih}
\affil[1]{Department of Radiology, Weill Cornell Medicine, New York, NY, USA}
\affil[2]{Department of Population Health Sciences, Weill Cornell Medicine, New York, NY, USA}
\affil[3]{Department of Radiology, Thomas Jefferson University, Philadelphia, PA, USA}
\affil[4]{Department of Medical Imaging, St. Michael’s Hospital/Unity Health Toronto, University of Toronto, Toronto, ON, Canada}
\affil[5]{Department of Radiology and Biomedical Imaging; Division of Clinical Informatics and Digital Transformation, Department of Medicine, University of California, San Francisco, CA, USA}
\affil[6]{Department of Radiology, Ohio State University Wexner Medical Center, OH, USA}
\affil[7]{Diagnostics Institute, Cleveland Clinic Foundation, Cleveland, OH, USA}
\affil[8]{Hospital Israelita Albert Einstein, Av. Albert Einstein, 627, São Paulo 05652, Brazil}
\affil[9]{Radiological Society of North America, Oak Brook, IL, USA}
\affil[10]{Department of Radiology, Brigham and Women's Hospital, Harvard Medical School, Boston, MA, USA}
\affil[11]{Department of Radiology, Perelman School of Medicine at the University of Pennsylvania, Philadelphia, PA, USA}
\affil[12]{Department of Diagnostic and Interventional Radiology, Klinikum rechts der Isar, Technical University Munich, Munich, Germany}
\affil[13]{Department of Radiology, Johns Hopkins University School of Medicine, Baltimore, MD, USA}
\affil[14]{Trillium Health Partners, Department of Medical Imaging, Faculty of Medicine, University of Toronto}
\affil[15]{Department of Materials Science and Engineering, Faculty of Engineering, University of Toronto}
\affil[16]{Joint Department of Medical Imaging, Toronto General Hospital, University of Toronto, Toronto, ON, Canada}
\affil[17]{St. Joseph’s Health Care London, Western University, London, ON}
\affil[18]{Department of Radiology and Imaging Sciences, Emory University School of Medicine, Atlanta, GA, USA}
\affil[19]{Department of Radiology, UT Southwestern Medical Center, Dallas, TX, USA}
\affil[*]{Corresponding author(s). Email(s): \url{george@cornellradiology.org}}

\begin{document}

\maketitle

\newpage

\begin{abstract}
\textbf{Background}:
Multimodal large language models have demonstrated comparable performance to that of radiology trainees on multiple-choice board-style exams. However, to develop clinically useful multimodal LLM tools, high-quality benchmarks curated by domain experts are essential.

\textbf{Purpose}:
To curate released and holdout datasets of 100 chest radiographic studies each and propose an artificial intelligence (AI)-assisted expert labeling procedure to allow radiologists to label studies more efficiently.

\textbf{Materials and Methods}:
A total of 13,735 deidentified chest radiographs and their corresponding reports from the MIDRC were used. GPT-4o extracted abnormal findings from the reports, which were then mapped to 12 benchmark labels with a locally hosted LLM (Phi-4-Reasoning). From these studies, 1,000 were sampled on the basis of the AI-suggested benchmark labels for expert review; the sampling algorithm ensured that the selected studies were clinically relevant and captured a range of difficulty levels. Seventeen chest radiologists participated, and they marked ``Agree all", ``Agree mostly'' or ``Disagree'' to indicate their assessment of the correctness of the LLM suggested labels. Each chest radiograph was evaluated by three experts. Of these, at least two radiologists selected ``Agree All'' for 381 radiographs. From this set, 200 were selected, prioritizing those with less common or multiple finding labels, and divided into 100 released radiographs and 100 reserved as the holdout dataset. The holdout dataset is used exclusively by RSNA to independently evaluate different models.

\textbf{Results}:
A benchmark of 200 studies (100 released and 100 reserved as holdouts) was developed, each containing one to four benchmark labels. Cohen's $\kappa$ for the agreement categories among experts was 0.622 (95\% CI 0.590, 0.651). At the individual condition (radiographic abnormality) level, other than airspace opacity ($\kappa$ 0.484 95\% CI [0.440, 0.524]), most conditions had a Cohen's $\kappa$ above 0.75 (range, 0.744--0.809) among experts.

\textbf{Conclusion}

A benchmark of 200 chest radiographic studies with 12 benchmark labels was created and made publicly available (\url{https://imaging.rsna.org}), with each chest radiograph verified by three radiologists. In addition, an AI-assisted labeling procedure was developed to help radiologists label at scale, minimize unnecessary omissions, and support a semicollaborative environment.
\end{abstract}


\section{Introduction}\label{introduction}

Multimodal large language models have achieved performance comparable to that of radiology trainees on board-style exams~\cite{bhayana2023performance}. However, it is still unclear how effective these models are in clinical settings, particularly in regard to abnormality detection on medical images. To properly evaluate these models, there is a critical need for high-quality benchmarks, curated by experts and grounded in complex, real-world scenarios. Such benchmarks can help provide an accurate and meaningful assessment of their performance, which may differ depending on the clinical presentation and disease. This benchmark was developed by the RSNA AI Committee Large Language Model (LLM) Workgroup. Chest radiographs were selected for the initial multimodal LLM benchmark because they are among the most widely performed imaging studies worldwide.

While several datasets and benchmarks have been published in the field of radiology, several limitations remain. First, most datasets (e.g., NIH-CXR~\cite{wang2017chestx-ray8}, MIMIC-CXR-JPG~\cite{johnson2019mimic-cxr-jpg}, RadGraph~\cite{jain2021radgraph}, and CheXpert~\cite{irvin2019chexpert}) have applied natural language processing (NLP) to extract labels from radiology reports alone and lack information derived from direct image interpretation. Second, past annotation efforts typically involved only a small number of radiologists rather than a broader pool of experts, reducing representativeness and increasing the potential for individual biases in the labeling process~\cite{majkowska2020chest}. Additionally, existing datasets available for use consist of downsampled images in JPEG or PNG rather than DICOM format, resulting in information loss.

To address these limitations, the RSNA AI Committee Large Language Model (LLM) Workgroup presents the RSNA LLM Benchmark Dataset: Radiologist Evaluation and Validation Enhanced by AI Labels (REVEAL-CXR), which offers several major contributions. A curated dataset annotated by 17 board-certified cardiothoracic radiologists from 10 institutions across 4 countries is presented, encouraging collaboration and reducing individual biases. An LLM-assisted workflow was implemented to support radiologists in reviewing images more efficiently by extracting labels from radiology reports, thereby minimizing missed diagnoses and facilitating collaborative discussion. The released dataset focuses on cardiothoracic abnormalities identified on chest radiographs, with diagnostic labels determined through radiologist interpretations of the images. Majority voting was used to achieve a consensus among the radiologists. Users can use the benchmark to evaluate their models. Although the sample size is not large, it focuses on rare conditions and complex cases involving multiple diseases, which are the most informative data points for model evaluation.

\section{Materials and Methods}\label{materials-and-methods}

\subsection{Data collection and initial label generation}\label{data-collection-and-initial-label-generation}

REVEAL-CXR uses deidentified chest radiographs and associated reports from the Medical Imaging and Data Resource Center (MIDRC), a registry of deidentified examinations and corresponding reports collected under IRB oversight beginning in 2020. As such, this study was exempt from the need for institutional review board approval. The chest radiographs and reports used in the development of the benchmark have not been previously released to the public (\url{https://www.midrc.org/}). For conciseness, we will use ``study'' as a shorthand for ``chest radiograph study'' throughout this work and use the terms interchangeably.

To ensure that this benchmark reflects clinically relevant and widely recognized findings on chest radiography, a multidisciplinary and multinational committee of radiologists and data scientists was convened. Using a formal consensus process including literature reviews, iterative discussions, and voting, a set of 12 benchmark labels was identified that represent commonly encountered findings on chest radiographs. This collaborative and cross-institutional approach ensured that the selected labels were both clinically meaningful and broadly applicable across diverse healthcare settings. The final set of labels included airspace opacity, aneurysm, cardiomegaly, chronic obstructive pulmonary disease (COPD), hiatal hernia, interstitial opacity, lung mass, lung nodule, lymphadenopathy, pleural effusion, pleural thickening, and pneumothorax. The mapping of these labels to RadLex and SNOMED is provided in \textbf{Extended Data Table~\ref{stab:mapping}}. Each study is allowed to have multiple benchmark labels.

\begin{figure}[t]
\centering
\includegraphics[width=0.3\linewidth]{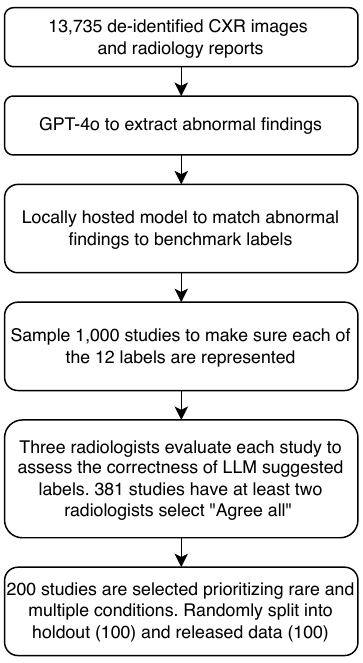}
\caption{Process for generating candidate findings from the original radiology reports, assisted by LLMs.}
\label{fig:process}
\end{figure}

The MIDRC provided access to an unreleased set of 13,735 deidentified chest radiograph studies paired with their original corresponding reports. LLMs were used to estimate the distribution of potential abnormalities (\textbf{Figure \ref{fig:process}}). This process involved two steps. First, a prompt was provided to a frontier LLM (GPT-4o 2024-08-01-preview~\cite{openai2023gpt-4}) (\textbf{Extended Data Box \ref{prompt:gpt4}}) to extract abnormal findings from the radiology reports; the instructions in the prompt are detailed in the Supplementary Materials. Second, a locally hosted large-language model (Phi-4-Reasoning~\cite{abdin2024phi-4}) was used to map the extracted findings to the predefined set of 12 benchmark labels. This mapping process uses the guided decoding function of the virtual LLM~\cite{kwon2023efficient} framework, which limits the output of the model to a structured format (\textbf{Extended Data Box \ref{box:schema}} and \textbf{Extended Data Box \ref{prompt:transform}}).

After generating the initial label predictions, a stratified sampling algorithm was applied to select studies with the benchmark labels identified by the Phi-4 model. The sample sizes for each stratum, that is, the number of labels per study, are as follows: 300 studies with one label, 300 with two labels, 319 with three labels, 76 with four labels, 4 with five labels, and 1 with six labels. This algorithm ensured that each study selected contained at least one of the 12 benchmark labels as specified by the LLM-generated annotations. Ultimately, a candidate dataset of 1,000 chest radiographs from different patients was created, selected for its clinical relevance and representing a range of difficulty levels.

\subsection{Data annotation}\label{data-annotation}

A multinational team of 17 fellowship-trained cardiothoracic radiologists was recruited from 10 institutions across 4 countries to adjudicate the labels extracted by the LLM for each imaging study. The radiologists' experience ranged as follows: G.B (21 years), D.D (15 years), N.S (6 years), F.S (8 years), J.B.A (15 years), E.N (20 years), G.S (26 years), L.C.A (9 years) D.A.T (25 years), K.H (11 years), L.G (6 years), H.S (15 years), A.P (2 years), A.I (21 years), M.C (7 years), G.T (17 years), and F.J (5 years). A web-based annotation platform, which shares features of a picture archiving and communication system (PACS), was used for this task.

\begin{figure}[t]
\centering
\includegraphics[width=\linewidth]{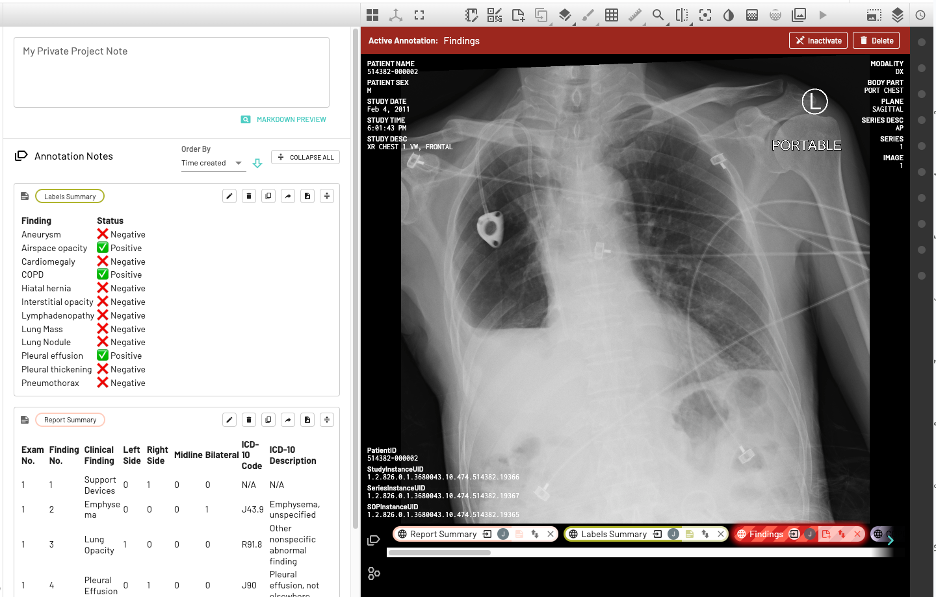}
\caption{The web-based platform where radiologists indicated \textit{Agree}, \textit{Agree Mostly}, or \textit{Disagree} with the label set extracted by an LLM from the original radiology report.}
\label{fig:platform}
\end{figure}

The radiologists were presented with the chest radiographs and a summary of the LLM-extracted labels (\textbf{Figure \ref{fig:platform}}) but not the original radiology reports. For each study, the radiologists were instructed to review the images and assess the suggested labels by selecting one of three options: ``Agree All,'' ``Agree Mostly,'' or ``Disagree.'' If the choice was not ``Agree All,'' an optional comment field allowed them to provide feedback on their choice. The radiologists were instructed to select ``Agree Mostly'' when the suggested labels were plausible, even in cases where a finding was subtle, equivocal, or potentially subject to interpretation (e.g., a faint pulmonary nodule). Each study was independently reviewed by three radiologists. The annotation platform was operated in ``crowd sourcing'' mode, allowing annotators to select the number of studies they wished to review. The studies were presented to the radiologists on the platform in random order. To promote diversity in the reviews and avoid overrepresentation from a small subset of reviewers, the radiologists were limited to reviewing no more than 300 studies. The radiologists were blinded to the assessments of their peers.

\subsection{Dataset curation}\label{dataset-curation}

To ensure that the final dataset was of sufficient quality, only studies with at least two ``Agree All'' ratings out of the three radiologists' ratings were considered, resulting in a total of 381 studies. Next, to maximize the utility of the dataset, all studies containing less common findings (COPD, pleural thickening, lung nodule, lung mass, aortic aneurysm, pneumothorax, hiatal hernia, lymphadenopathy, and interstitial opacity) were first retained, yielding a dataset of 123 studies. These studies were then randomly split into a released dataset and a holdout dataset. The remaining studies were then ranked by the number of findings; an additional 77 studies with the highest label counts were selected and randomly split. In the end, both the released and holdout datasets contained 100 studies each. The level of agreement among experts (whether all selected ``agree all'' or only two did) is also included in the dataset.

\section{Statistical methods}\label{statistical-methods}

To assess whether the released and holdout subsets of the REVEAL-CXR dataset are comparable in imaging acquisition properties, we evaluated key DICOM characteristics -- such as manufacturer, model name, detector type, view position, KVP, exposure, and pixel spacing -- and compared their distributions using Pearson's Chi-square test. The $\chi^2$ test is selected because the variables are categorical or discretized, and the goal is to determine whether the two subsets differ in their distributions. All characteristics showed p-values above 0.05 (e.g., Manufacturer p=0.057, Detector Type p=0.726), indicating no significant differences. Overall, the acquisition-related distributions are well balanced between released and holdout data.

Cohen's $\kappa$ was used to measure the agreement level between two categorical ratings. $\kappa$ compares the observed agreement to that expected from the raters' marginal label frequencies, with values closer to 1 indicating stronger concordance. For each $\kappa$ estimate, 1,000 bootstrap resamples were used to compute 95\% confidence intervals. Majority vote served as the reference standard in two settings: (1) comparing each radiologist's binary ``Agree'' versus ``Disagree'' rating to the group consensus, and (2) constructing a study-level findings list when radiologists adjusted the LLM-suggested labels. $\kappa$ was then calculated between each radiologist and this majority-voted reference to summarize interrater reliability at both the study and finding levels.

\section{Results}\label{results}

\subsection{\texorpdfstring{Descriptive statistics }{Descriptive statistics }}\label{descriptive-statistics}

The acquisition characteristics of both the holdout and released datasets, including manufacturer, model, detector type, view position, etc, are summarized in \textbf{Extended Data Table \ref{stab:acquisition}}. The corresponding p-values from Chi-square tests assessing balance between the two datasets are also reported. All p-values exceed 0.05, indicating no significant differences. The demographic characteristics of the final released and holdout datasets are detailed in \textbf{Table \ref{tab:data}}.
\begin{table}[t]
\centering
\caption{Demographic and patient characteristics of the benchmark dataset.}
\label{tab:data}
\begin{tabular}[]{lrr}
\toprule
& \textbf{Released} & \textbf{Holdout} \\
\midrule
\textbf{Total} & 100 & 100 \\
\midrule
\textbf{Age} & & \\
~~~~80+ & 23 & 23 \\
~~~~70-79 & 26 & 29 \\
~~~~60-69 & 22 & 20 \\
~~~~50-59 & 15 & 16 \\
~~~~40-49 & 13 & 5 \\
~~~~0-39 & 1 & 4 \\
~~~~Unknown & 0 & 3 \\
\midrule
\textbf{Sex} & & \\
~~~~Female & 50 & 49 \\
~~~~Male & 50 & 48 \\
~~~~Unknown & 0 & 3 \\
\midrule
\textbf{Race} & & \\
~~~~Asian & 6 & 5 \\
~~~~African American & 14 & 20 \\
~~~~White & 46 & 34 \\
~~~~Other & 3 & 2 \\
~~~~Unknown & 31 & 39 \\
\midrule
\textbf{Patient class} & & \\
~~~~Inpatient & 89 & 80 \\
~~~~Outpatient & 11 & 17 \\
~~~~Unknown & 0 & 3 \\
\bottomrule
\end{tabular}
\end{table}

\begin{figure}[t]
\centering
\includegraphics[width=.67\linewidth]{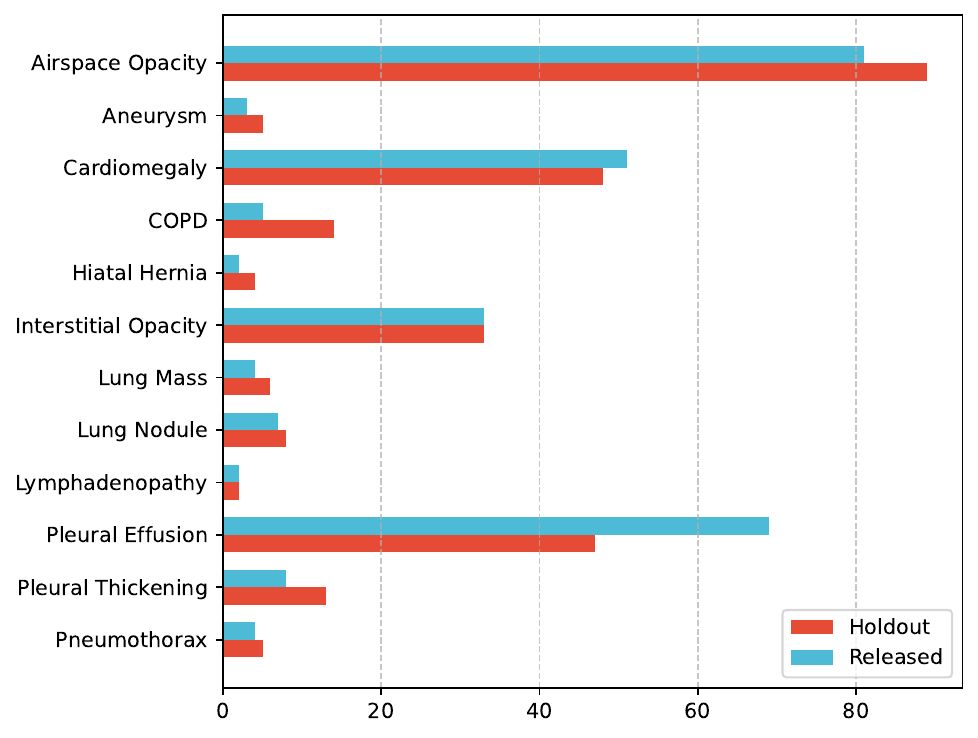}
\caption{Prevalence of radiologic findings across the released and holdout benchmark datasets. Distribution of the findings in the released and holdout datasets.}
\label{fig:prevalence}
\end{figure}

\textbf{Figure \ref{fig:prevalence}} presents the frequency of each final annotated finding. Airspace opacity was the most prevalent finding in the dataset, followed by pleural effusion, cardiomegaly, and interstitial opacity. The overall distribution of the findings exhibited a strong, long-tailed pattern, with less common findings, such as hiatal hernia and lymphadenopathy, appearing in fewer than ten cases.

\begin{figure}[t]
\centering
\includegraphics[width=.67\linewidth]{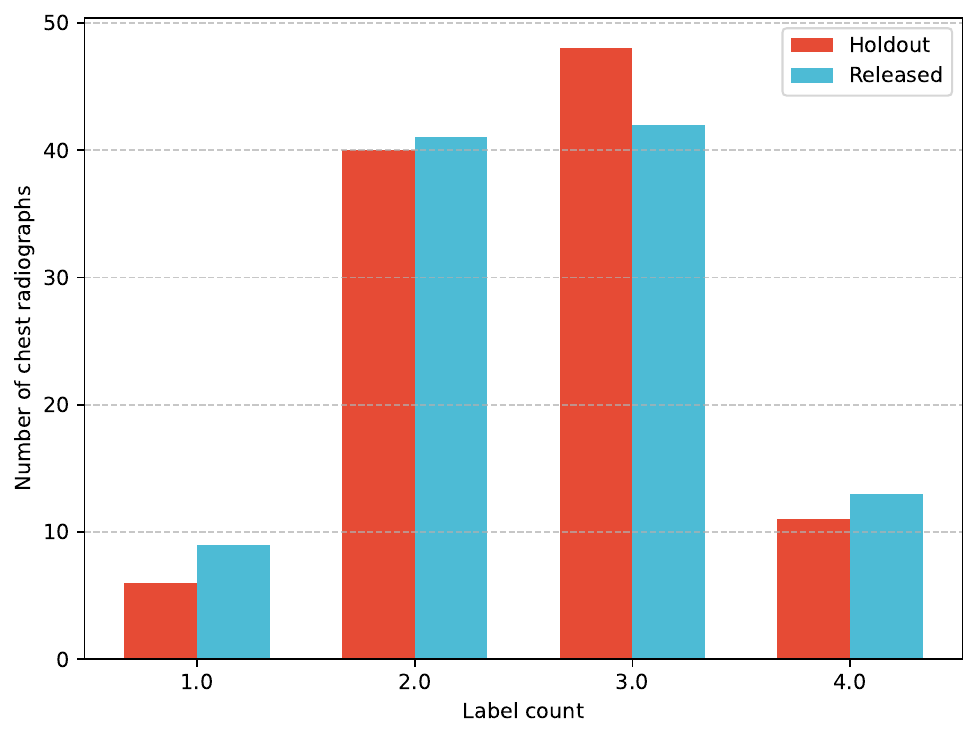}
\caption{Distribution of the number of findings per chest radiograph in the released and holdout datasets.}
\label{fig:distribution}
\end{figure}

\textbf{Figure \ref{fig:distribution}} shows the distribution of final finding counts per study. Because studies with more findings were oversampled when selecting the final datasets, the label distribution shows higher finding counts than does the sampling distribution described in the Methods section.

\subsection{Interannotator agreement}\label{interannotator-agreement}

We first assessed the agreement between radiologist annotators regarding the labels 'agree all,\textquotesingle{} 'agree mostly,\textquotesingle{} and 'disagree. Because the radiologists did not consistently distinguish between ``Disagree'' and ``Agree Mostly,'' these responses were combined into a single ``Disagree'' category. On this binary scale, Cohen's $\kappa$ between individual radiologists and the majority vote (i.e., at least 2 raters selecting the same category) was 0.622 (1,000 bootstrapped resamples, 95\% CI: 0.590, 0.651), reflecting substantial agreement according to the Landis \& Koch scale. The lower end of the confidence interval, 0.590, falls in the 'moderate range\textquotesingle, which is a more conservative estimation. The relatively low agreement level is primarily attributable to the specific condition of airspace opacity, which is discussed later. Furthermore, this disagreement reflects the inherent complexity of real-world radiographs. These difficult cases are valuable for evaluating LLMs, as hard examples reveal model limitations more effectively. Regarding the extent to which radiologists agree with the labels suggested by LLMs, in 619/1,000 cases (61.9\%), the majority of the votes fell into the ``Disagree'' category, indicating that the radiologists' labels often diverged from those suggested by the LLM.

Given the frequent divergence from the LLM labels, next, the interrater agreement among radiologists was examined for individual findings (Extended Data Table \ref{stab:cohen}). When a radiologist chose ``Agree All'', the original LLM-suggested list was retained; otherwise, the candidate findings were adjusted on the basis of the annotator's notes (Extended Data Box \ref{prompt:revised}). Because not all radiologists provided notes when they disagreed, we constructed a majority-voted findings list for 681 studies. Cohen's $\kappa$ was then calculated between each radiologist's annotation and this majority list. For most findings, the agreement was relatively high: nine findings had a $\kappa$ value greater than 0.7, with the highest calculated for hiatal hernia ($\kappa$ = 0.809, 95\% CI [0.688, 0.902]), indicating substantial consistency across the radiologists. The notable exception was airspace opacity ($\kappa$ = 0.484, 95\% CI [0.440, 0.524]), which is consistent with previous reports of low interrater agreement for this finding~\cite{Loeb2006-jf, voigt2021interobserver, albaum1996interobserver, makhnevich2019clinical}.

\section{Discussion}\label{discussion}

In this work, we developed a benchmark specifically focused on cardiothoracic findings on chest radiographs. We also presented a process that uses an LLM to assist radiologists with efficient labeling and to streamline data curation for the creation of future benchmarks. The RSNA AI Committee has already created several labeled datasets for ML challenges (\url{https://imaging.rsna.org}), which were annotated de novo (no suggested labels). We believe that with the help of LLMs to parse these reports to provide initial labels, radiologists can validate these annotations more efficiently than by creating annotations without this step. Additionally, leveraging LLMs in this process should not compromise accuracy, as each study is still verified by three radiologists. This framework is particularly valuable for creating larger-scale benchmarks in the future. Furthermore, the benchmark dataset was curated in collaboration with a large group of subspecialized radiologists from across the world, which is unusual for most existing benchmarks. The original images, provided in DICOM format, were used to maintain the highest quality and are available in the final benchmark dataset.

The benchmark produced with this method has several limitations. First, the process for mapping the findings to the twelve benchmark labels relies only on the findings extracted by the first LLM rather than the entire report. Second, we used a relatively small model (Phi-4-Reasoning) for the mapping step; while this approach is faster, its performance may not match that of larger models such as GPT-4o. Therefore, the mapping performance may be suboptimal. Third, we lacked access to comprehensive patient clinical conditions and demographic characteristics, which are often crucial for achieving a more accurate diagnosis. Finally, more granular criteria for annotation could be implemented. Upon reviewing the annotations, we found that the radiologists did not systematically distinguish between ``Disagree'' and ``Agree Mostly'', which required us to treat these responses as equivalent during postprocessing.

As large multimodal LLMs become more prevalent and increasingly accepted by patients and healthcare providers, even prior to formal clinical validation and potentially outside of controlled clinical environments, robust benchmarking will become essential. These benchmarks may offer valuable insights into the expected performance of emerging multimodal LLMs in real-world scenarios. In future studies, we plan to expand our efforts by developing additional multimodal benchmarks encompassing other medical imaging modalities, including computed tomography (CT), magnetic resonance imaging (MRI), ultrasonography, and others.

\section*{Funding Sources}

This work was supported by the National Institutes of Health (NIH) under grant numbers R01CA289249 and 75N92020D00021, U.S. National Science Foundation (NSF) under grant numbers NSF CAREER 2145640.



\setlength{\bibsep}{3pt plus 0.3ex}
\bibliographystyle{unsrtnat}
\bibliography{paperpile}

\newpage
\appendix
\setcounter{table}{0}
\setcounter{figure}{0}
\renewcommand\figurename{Figure}
\renewcommand\tablename{Table}
\renewcommand{\thefigure}{S\arabic{figure}}
\renewcommand{\thetable}{S\arabic{table}}

\section*{Supplementary data}
\label{sec:appendix}


\begin{longtblr}[
    caption = {Mapping of disease labels to RadLex and SNOMED terms},
    label = {stab:mapping},
]{
    colspec = {l>{\raggedright\arraybackslash}X>{\raggedright\arraybackslash}X>{\raggedright\arraybackslash}X>{\raggedright\arraybackslash}X},
    width=\textwidth,
    cells = {font = \small},
    rowhead = 1,
    row{odd} = {gray!10},
    row{1} = {white}, 
    hline{1,2,Z} = {0.08em, black}
}
Finding / Label & Operational Definition (Radiologic) & RadLex Term (RID) & SNOMED CT Concept ID & LOINC Code (Name)  \\
Aneurysm & Focal dilation of a vessel (artery or aorta) $>50\%$ of its normal diameter. & RID28677 – Aneurysm (morphologic abnormality) & 233985008 – Aneurysm (disorder) & LP38233-3 – Aneurysm [Imaging observation]  \\
Airspace opacity & Increased parenchymal attenuation obscuring vascular markings due to alveolar filling (fluid, pus, blood, cells). & RID28663 – Airspace opacity (finding) & 301857004 – Air space opacity (finding) & LP38249-9 – Airspace opacity [Imaging observation]  \\
Cardiomegaly & Cardiac silhouette enlargement (cardiothoracic ratio $> 0.5$ on PA film). & RID35639 – Cardiomegaly (finding) & 8186001 – Cardiomegaly (disorder) & LP38234-1 – Cardiomegaly [Imaging observation]  \\
COPD & Chronic airflow limitation with radiographic signs such as hyperinflated lungs, flattened diaphragm, and decreased vascular markings. & RID43230 – Chronic obstructive pulmonary disease & 13645005 – Chronic obstructive lung disease (disorder) & LP38235-8 – Chronic obstructive pulmonary disease [Imaging observation]  \\
Hiatal hernia & Protrusion of stomach through the esophageal hiatus into the thorax. & RID5671 – Hiatal hernia & 39839004 – Hiatus hernia (disorder) & LP38237-4 – Hiatal hernia [Imaging observation]  \\
Interstitial opacity & Fine or coarse linear, reticular, or reticulonodular opacities due to interstitial thickening or fibrosis. & RID28664 – Interstitial opacity (finding) & 301858009 – Interstitial opacity (finding) & LP38250-7 – Interstitial opacity [Imaging observation]  \\
Lymphadenopathy & Enlargement of lymph nodes, usually $>1$ cm short-axis, may be hilar or mediastinal. & RID5694 – Lymphadenopathy & 30746006 – Lymphadenopathy (disorder) & LP38238-2 – Lymphadenopathy [Imaging observation]  \\
Lung mass & Focal pulmonary opacity $>3$ cm in diameter. & RID43235 – Lung mass & 126713003 – Mass of lung (finding) & LP38248-1 – Mass [Imaging observation]  \\
Lung nodule & Rounded or irregular pulmonary opacity $\leq 3$ cm in diameter. & RID12780 – Pulmonary nodule & 39607008 – Lung nodule (finding) & LP38246-5 – Nodule [Imaging observation]  \\
\end{longtblr}

\newpage


\begin{longtblr}[
    caption = {Key acquisition characteristics of holdout and released datasets},
    label = {stab:acquisition},
]{
    colspec = {Xrrr},
    width=.85\textwidth,
    cells = {font = \small},
    rowhead = 1,
    row{odd} = {gray!10},
    row{1} = {white}, 
    hline{1,2,Z} = {0.08em, black}
}
& p-value & holdout & released  \\ 
\textbf{Manufacturer} & 0.057\\
CARESTREAM && 1.0 & 0.0  \\ 
CARESTREAM HEALTH && 49.0 & 37.0  \\ 
Carestream && 5.0 & 10.0  \\ 
Carestream Health && 35.0 & 39.0  \\ 
GE Healthcare && 8.0 & 5.0  \\ 
GE MEDICAL SYSTEMS && 2.0 & 0.0  \\ 
Imaging Dynamics Company Ltd. && 0.0 & 4.0  \\ 
KODAK && 4.0 & 1.0  \\ 
KONICA MINOLTA && 1.0 & 1.0\\
Philips Medical Systems && 7.0 & 2.0\\
\hline
\textbf{Manufacturer Model Name} & 0.066\\
CS-7 && 1.0 & 1.0  \\ 
Definium 5000 && 5.0 & 11.0  \\ 
DigitalDiagnost && 7.0 & 2.0  \\ 
Discovery XR656 && 8.0 & 5.0  \\ 
DR 7500 && 2.0 & 0.0  \\ 
DRX-1 && 1.0 & 2.0  \\ 
DRX-EVOLUTION && 2.0 & 1.0  \\ 
DRX-REVOLUTION && 50.0 & 36.0  \\ 
DRX-Revolution && 39.0 & 48.0  \\ 
Revolution XRd ADS\_28.4 && 2.0 & 0.0\\
X3 && 0.0 & 4.0\\
\hline
\textbf{Detector Type} & 0.726\\
DIRECT && 78 & 77\\
SCINTILLATOR && 22 & 18\\
\hline
\textbf{View Position} & 0.115\\
AP && 96.0 & 88  \\ 
LATERAL && 0.0 & 2  \\ 
LL && 4.0 & 8  \\ 
PA && 16.0 & 9  \\ 
POSTERO\_ANTERIOR && 0.0 & 2  \\ 
\hline
\textbf{Distance Source To Detector} & 0.426\\
0.0 && 8.0 & 10.0  \\ 
12.4 && 1.0 & 0.0  \\ 
18.4 && 1.0 & 0.0  \\ 
1000.0 && 9.0 & 13.0  \\ 
1800.0 && 2.0 & 0.0  \\ 
1810.0 && 1.0 & 0.0  \\ 
1811.0 && 0.0 & 1.0  \\ 
1817.0 && 2.0 & 0.0  \\ 
1828.0 && 4.0 & 3.0  \\ 
1829.0 && 2.0 & 1.0  \\ 
1830.0 && 1.0 & 0.0\\
1885.0 && 1.0 & 0.0\\
\hline
\textbf{Distance Source To Patient} & 0.152\\
946.0 && 5.0 & 11.0  \\ 
994.0 && 4.0 & 2.0  \\ 
1751.0 && 2.0 & 0.0  \\ 
1756.0 && 1.0 & 0.0  \\ 
1757.0 && 0.0 & 1.0  \\ 
1763.0 && 2.0 & 0.0  \\ 
1775.0 && 2.0 & 0.0  \\ 
1776.0 && 5.0 & 3.0  \\ 
1831.0 && 1.0 & 0.0  \\ 
\hline
\textbf{KVP} & 0.341\\
83.0 && 0.0 & 1.0  \\ 
85.0 && 36.0 & 33.0  \\ 
87.0 && 1.0 & 0.0  \\ 
88.0 && 1.0 & 0.0  \\ 
89.0 && 0.0 & 2.0  \\ 
90.0 && 2.0 & 2.0  \\ 
91.0 && 0.0 & 1.0  \\ 
95.0 && 5.0 & 2.0  \\ 
100.0 && 1.0 & 3.0  \\ 
104.0 && 1.0 & 0.0  \\ 
105.0 && 1.0 & 0.0  \\ 
106.0 && 1.0 & 0.0  \\ 
107.0 && 0.0 & 1.0  \\ 
110.0 && 46.0 & 33.0  \\ 
112.0 && 0.0 & 1.0  \\ 
115.0 && 6.0 & 5.0  \\ 
116.0 && 0.0 & 1.0  \\ 
120.0 && 10.0 & 20.0  \\ 
125.0 && 3.0 & 2.0  \\ 
140.0 && 1.0 & 0.0  \\  
\hline
\textbf{Exposure Inm As} & 0.447\\
1.1 && 0.0 & 1.0  \\ 
1.6 && 14.0 & 15.0  \\ 
1.7 && 1.0 & 1.0  \\ 
1.8 && 1.0 & 0.0  \\ 
2.0 && 12.0 & 9.0  \\ 
2.2 && 0.0 & 2.0  \\ 
2.3 && 1.0 & 0.0  \\ 
2.5 && 0.0 & 5.0  \\ 
2.8 && 1.0 & 5.0  \\ 
3.1 && 3.0 & 4.0 \\ 
3.2 && 11.0 & 11.0  \\ 
3.6 && 1.0 & 0.0  \\ 
3.9 && 1.0 & 0.0  \\ 
4.0 && 3.0 & 5.0  \\ 
4.5 && 2.0 & 1.0  \\ 
5.0 && 0.0 & 2.0  \\ 
5.6 && 1.0 & 1.0  \\ 
6.3 && 2.0 & 1.0  \\ 
7.1 && 1.0 & 2.0  \\ 
8.0 && 1.0 & 0.0  \\ 
10.1 && 0.0 & 1.0  \\ 
12.5 && 1.0 & 0.0  \\ 
19.8 && 0.0 & 1.0  \\ 
25.0 && 1.0 & 0.0  \\ 
\hline
\textbf{Pixel Spacing} & 0.220\\
0.139$\backslash$0.139 && 92.0 & 87.0  \\ 
0.143$\backslash$0.143 && 2.0 & 0.0  \\ 
0.144$\backslash$0.144 && 0.0 & 4.0  \\ 
0.1488636604775$\backslash$0.1488636604775 && 1.0 & 0.0  \\ 
0.1488896174863$\backslash$0.1488896174863 && 1.0 & 0.0  \\ 
0.1488901038819$\backslash$0.1488901038819 && 2.0 & 0.0  \\ 
0.1488959823886$\backslash$0.1488959823886 && 2.0 & 0.0  \\ 
0.1488989508559$\backslash$0.1488989508559 && 0.0 & 1.0  \\ 
0.1488994475138$\backslash$0.1488994475138 && 1.0 & 0.0  \\ 
0.14948$\backslash$0.14948 && 0.0 & 1.0  \\ 
0.175$\backslash$0.175 && 1.0 & 1.0  \\ 
0.194311$\backslash$0.194311 && 4.0 & 3.0  \\ 
0.1988$\backslash$0.1988 && 4.0 & 2.0  \\ 
\end{longtblr}

\newpage

\begin{table}[!ht]
\centering
\caption{Cohen's Kappa at disease level and prevalence in the 1000 sampled studies}
\label{stab:cohen}
\begin{tabular}{lr@{~}rr@{~}rr}
\toprule
Disease & \multicolumn{2}{r}{Cohen's Kappa} & \multicolumn{2}{r}{Gwet's AC1} & \makecell[br]{Positive ratio for\\final annotation}\\ 
\midrule
airspace opacity & 0.484 & [0.440, 0.524] & 0.501 & [0.445, 0.556] & 0.846  \\ 
pleural effusion & 0.756 & [0.727, 0.784] & 0.667 & [0.617, 0.716] & 0.455  \\ 
cardiomegaly & 0.759 & [0.729, 0.789] & 0.633 & [0.582, 0.685] & 0.460  \\ 
interstitial opacity & 0.746 & [0.710, 0.780] & 0.759 & [0.716, 0.801] & 0.230  \\ 
copd & 0.759 & [0.689, 0.821] & 0.955 & [0.939, 0.970] & 0.055  \\ 
pleural thickening & 0.744 & [0.669, 0.811] & 0.948 & [0.932, 0.965] & 0.048  \\ 
lung nodule & 0.709 & [0.626, 0.781] & 0.959 & [0.945, 0.974] & 0.049  \\ 
lung mass & 0.801 & [0.714, 0.875] & 0.979 & [0.969, 0.989] & 0.033  \\ 
aneurysm & 0.651 & [0.536, 0.758] & 0.965 & [0.952, 0.978] & 0.025  \\ 
pneumothorax & 0.753 & [0.617, 0.855] & 0.982 & [0.972, 0.991] & 0.018  \\ 
hiatal hernia & 0.809 & [0.688, 0.902] & 0.988 & [0.980, 0.995] & 0.017  \\ 
lymphadenopathy & 0.634 & [0.454, 0.775] & 0.984 & [0.976, 0.993] & 0.013  \\ 
\bottomrule
\end{tabular}
\end{table}

\newpage

\begin{table}[!ht]
\centering
\caption{Studies finished per radiologist}
\label{stab:studies}
\begin{tabular}{lr}
\toprule
Radiologist & Studies finished  \\ 
\midrule
N.S & 300  \\ 
G.B & 300  \\ 
D.D & 300  \\ 
F.S & 273  \\ 
J.A & 229  \\ 
E.N & 222  \\ 
G.S & 213  \\ 
D.T & 203  \\ 
K.H & 177  \\ 
L.G & 151  \\ 
H.S & 140  \\ 
A.P & 133  \\ 
A.I & 121  \\ 
L.A & 94  \\ 
M.C & 57  \\ 
G.T & 53  \\ 
F.J & 34  \\ 
\bottomrule
\end{tabular}
\end{table}

\newpage


\begin{pboxed}[Prompt for GPT4-o to extract clinical findings from radiology reports]
\label{prompt:gpt4}
\begin{lstlisting}[language=prompt]
Find all diseases in the report that can have an ICD-10 code, and provide a summary in a table format where the positive clinical conditions are 1 and the negative clinical conditions are 0. Designate each condition as left side, right side, midline, or bilateral. Provide an ICD-10 code and ICD-10 Description in separate columns for positive findings only or N/A if not applicable.

Table columns include: [Exam No., Finding No., Clinical Finding, Left Side, Right Side, Midline, Bilateral, Midline, ICD-10 Code, ICD-10 Description]

Additional instructions:

1. Normal findings should be excluded from each table
2. Group similar findings together where possible for each table
3. Create a table and also a code block highlighted CSV (without quotes)

Here is the full report:
{note}
\end{lstlisting}
\end{pboxed}

\newpage

\begin{pboxed}[Schema for transfer table to benchmark labels]
\label{box:schema}
\begin{lstlisting}[language=json]
benchmark_labels_schema = {
 "$schema": "https://json-schema.org/draft/2020-12/schema",
 "$id": "https://example.com/keyfindings.schema.json",
 "title": "KeyFindingsToLabels",
 "description": "Schema for mapping key findings to benchmark labels.",
 "type": "object",
 "properties": {
   "key_findings": {
     "type": "array",
     "items": {
       "type": "string"
     },
     "description": "List of key positive (abnormal) findings. If no abnormalities, the array should contain 'Normal exam'."
   },
   "benchmark_labels": {
     "type": "array",
     "items": {
       "type": "string",
       "enum": [
         "Aneurysm",
         "Airspace opacity",
         "Cardiomegaly",
         "COPD",
         "Hiatal hernia",
         "Interstitial opacity",
         "Lymphadenopathy",
         "Lung Mass",
         "Lung Nodule",
         "Pleural effusion",
         "Pleural thickening",
         "Pneumothorax",
         "None"
       ]
     },
     "minItems": 1,
     "description": "Zero or more labels from key_findings; otherwise return None."
   }
 },
 "required": ["key_findings", "benchmark_labels"],
 "additionalProperties": False
}
\end{lstlisting}
\end{pboxed}

\newpage

\begin{pboxed}[Prompt to transform table to benchmark labels]
\label{prompt:transform}
\begin{lstlisting}[language=prompt]
For each finding in the <Findings> table:

- Match finding to ONLY ONE corresponding label from <Labels> if possible
- Otherwise return 'None' for that finding
- Each finding in <Findings> table should return exactly ONE <Label>
- The number of Findings should EQUAL the number of corresponding Labels in the final output. It also equals the number of findings in the input table.
- The output is structured as JSON, with two keys 'key_findings' extracted from table and 'benchmark_labels' being your matched label

    <Labels>
    Aneurysm
    Airspace opacity
    Cardiomegaly
    COPD
    Hiatal hernia
    Interstitial opacity
    Lymphadenopathy
    Lung Mass
    Lung Nodule
    Pleural effusion
    Pleural thickening
    Pneumothorax
    </Labels>

    <Examples For Labels>
    Aneurysm = Aortic Aneurysm
    Airspace opacity = atelectasis, pneumonia, hemorrhage
    COPD = emphysema, hyperinflated lungs
    Lymphadenopathy = usually in hilar region

    Granuloma is a benign finding
    </Examples For Labels>

######## Here is the clinical findings input ########

```
{table}

######## Your output ########
\end{lstlisting}
\end{pboxed}

\newpage

\begin{pboxed}[Prompt for GPT4-o to suggest revised label based on radiologist comment]
\label{prompt:revised}
\begin{lstlisting}[language=prompt]
Another LLM is trying to predict disease from a radiology report. The diseases are from this list:
['Aneurysm', 'Airspace opacity', 'Cardiomegaly', 'COPD', 'Hiatal hernia', 'Interstitial opacity', 'Lymphadenopathy', 'Lung Mass', 'Lung Nodule', 'Pleural effusion', 'Pleural thickening','Pneumothorax']

Then we have a radiologist to evaluate its diagnosis. The radiologist will give a general comment (agree mostly, disagree) and note. I want you to give the final list based on the radiologist's comment.

#### Instructions:
1. Radiologist comment will only be 'disagree' or 'agree mostly'
2. The radiologist note is most informative. If the radiologist didn't give a meaningful note or there is no note, which can be very common due to system error. Return "NA"
3. Only return diseases that are certain. If the radiologist is not certain, do not include it as well.
4. Give your output in the first line, followed by your reasoning starting second line

#### Example 1:
* LLM Diagnosis: ['Airspace Opacity', 'Interstitial Opacity']
* Radiologist comment: Disagree.
* Radiologist note: No pneumonia
* Output
NA
Reason: The radiologist comment is not making sense since the LLM didn't mention pneumonia

#### Example 2:
* LLM Diagnosis: ['Airspace Opacity']
* Radiologist comment: Disagree
* Radiologist note: also cardiomegaly and pleural effusions
* Output:
[Airspace Opacity, Cardiomegaly, Pleural effusion]
Reason: Radiologist add those diseases to the list

#### Here is your input:
* LLM Diagnosis: {llm_labels}
* Radiologist comment: {annotation}
* Radiologist note: {note}
* Output:
\end{lstlisting}
\end{pboxed}

\end{document}